\documentclass[english]{article}
\usepackage[T1]{fontenc}
\usepackage[latin9]{inputenc}
\usepackage{geometry}
\usepackage{amsthm}
\usepackage{amsmath}
\usepackage{amssymb}
\usepackage{graphicx}

\makeatletter
  \theoremstyle{definition}
  \newtheorem{defn}{\protect\definitionname}
\theoremstyle{plain}
\newtheorem{thm}{\protect\theoremname}

\makeatother

\usepackage{babel}
  \providecommand{\definitionname}{Definition}
\providecommand{\theoremname}{Theorem}

\begin{document}

\title{Two Differentially Private Rating Collection Mechanisms for Recommender
Systems}

\author{Zheng Wenjie}
\maketitle
\begin{abstract}
We design two mechanisms for the recommender system to collect user
ratings. One is modified Laplace mechanism, and the other is randomized
response mechanism. We prove that they are both differentially private
and preserve the data utility.
\end{abstract}

\section{Introduction\label{sec:Introduction}}

Recommender Systems (RS) \cite{ricci2011introduction} are a kind
of system that seek to recommend to users what they are likely interested
in. Unlike search engines, the users do not need to type any keyword.
The RS's will learn their interest automatically. For instance, if
the user has just bought a numeric camera, the RS will recommend to
him some SD memory cards; if a user watches a lot of action movies,
the RS may suggest some other action movies to him. And this is the
typical behaviors which we observe universally in Netflix (movies),
Youtube (videos), Google Play (apps), Facebook (friends), Amazon (goods)
and other platforms today.

One may wonder how it works. Let us take Netflix as an example. Netflix
has a mechanism that allows every user to rate the movies they have
watched. Based on these ratings, Netflix builds a profile for each
user. And there should be quite a few methods to predict the user
preference on other movies that the user has yet seen. Readers can
learn more in Section~\ref{sec:Utility}. This is not the main topic
of this article.

The issue addressed in this article is whether user privacy is compromised
by the rating collection mechanism, and what we may do to prevent
it. In 2006, Netflix published 100~480~507 ratings that 480~189
users gave to 17~770 movies on the Internet to hold the \emph{Netflix
Prize} competition \cite{bennett2007netflix}. The data are anonymous.
However, in 2007, two researchers from the University of Texas, de-anonymized
some of the Netflix data by matching the data set with movie ratings
on the Internet Movie Database \cite{narayanan2008robust}. This aroused
big privacy concern. In 2009, four Netflix users filed a lawsuit against
Netflix. We see that this concern of privacy leak is real, and the
anonymization alone is not sufficient to prevent it.

Let us take a close look. There were actually two privacy leaks during
the procedure. Firstly, the users leaked their ratings to the service
provider Netflix. Then, Netflix leaked the ratings to the public.
Users made a big fuss on the second leak, but they overlooked the
fact that it was themselves who leaked the ratings to Netflix at the
first place. Usually, all legal companies will ask the users to sign
a user agreement, which authorizes the companies to collect user data
and to use them for certain purposes. However, almost no users will
ever read it. Anyway, if they do not agree, they will not be able
to use the service.

Hence, the goal of this article is to minimize the privacy leak but
still guaranteeing the functionality of the service provider. We will
achieve this goal by building differential privacy (DP) into the rating
collection mechanism. The concept of DP will be explained in detail
in Section~\ref{sec:Mechanisms}. The big idea is that the user ratings
are transformed through the rating collection mechanism, so that from
the output (transformed ratings), one cannot know for sure what the
input (original ratings) is. Of course, this kind of transform should
satisfy certain properties. After this transform, the service provider
can do whatever they want with the ratings without worrying about
privacy leak. They can either analyze it themselves, or subcontract
the work to a third party by giving them the data access. It is also
possible for Netflix to hold a second competition.

One trivial transform is to transform every rating to zero or pure
random number. This absolutely prevents any privacy leak, but it erases
all information contained in the data as well. Therefore, when we
build DP into the mechanism, we should be careful in order to preserve
as much information as possible in the data. We designed two mechanisms.
One is \emph{modified Laplace mechanism}, and the other is \emph{randomized
response mechanism} (Section~\ref{sec:Mechanisms}). We will show
that they preserve the utility of the ratings (Section~\ref{sec:Utility}).

\subsection*{Related work}

\cite{mcsherry2009differentially} also tries to bring DP to RS. Their
method is different from ours. Let $X^{i}$ be the original rating
set of the $i$-th user, $S$ be some aggregation statistic of ratings,
$\mathcal{A}$ be some algorithm to do data analysis, and $f$ be
some transform that guarantees DP. Their method can be summarized
as $\mathcal{A}\left(f\left(S\left(\otimes_{i=1}^{n}X^{i}\right)\right)\right)$
, while our method, with a little abuse of notation, can be summarized
as $\mathcal{A}\left(S\left(\otimes_{i=1}^{n}f\left(X^{i}\right)\right)\right)$. 

Note that we changed the position of the transform $f$. This modification
is of significant advantage. In their method, $f$ should be adapted
to each statistic $S$, and they can only use algorithms $\mathcal{A}$
relying on $S$. In our method, we can generalize it to $\mathcal{A}\left(\otimes_{i=1}^{n}f\left(X^{i}\right)\right)$,
which means that we can use more types of algorithms. Furthermore,
as long as $f$ is ``conjugate'' (i.e. $f$ does not change the
space where $X^{i}$ is in), all previous successful algorithms could
be seamlessly ``transplanted''. And we will illustrate in Section~\ref{sec:Utility}
that this transplantation is also seamless in theoretical guarantee.

When giving it a second thought, their method is neither privacy preserved
nor meaningful. According to their method, at the moment where users
transfer their ratings to the service provider so that it could calculate
the statistic $S$, user privacy has already leaked to the service
provider. Then, the service provider sends a differentially private
version of recommendation back to the user. But why the user bothers
to protect the privacy against himself?!

\section{Mechanisms\label{sec:Mechanisms}}

In this section, we will first introduce the concept of \emph{differential
privacy}. Then we will define the modified Laplace mechanism and randomized
response mechanism. Throughout this section, we consider the rating
vector of a single user: $x=\left(x_{1},x_{2},\ldots,x_{n}\right)$,
where $n$ is the number of items. Note that the components may have
missing values.

\subsection{Differentially private mechanism}

The concept of DP is not some entity lurking in the data, but it is
used to describe a certain type of data releasing mechanism. The original
idea is introduced in \cite{dwork2006calibrating}. Since, dozens
of formulations came out. We use the simplest formulation here.
\begin{defn}
Let $\epsilon$ be a positive value, a random application $\mathcal{M}:\mathfrak{R\longrightarrow\mathfrak{S}}$
is called $\epsilon$-differentially private mechanism if 
\[
\Pr\left(\mathcal{M}\left(y\right)\in S\right)\le\exp\left(\epsilon\right)\Pr\left(\mathcal{M}\left(z\right)\in S\right),
\]
 for any $y,z\in\mathfrak{R}$ and any $S\subset\mathfrak{S}$. 
\end{defn}
The idea is that the distributions produced by $y$ and $z$ are absolutely
continuous to each other with the multiplier $\exp\left(\epsilon\right)$.
With $\epsilon$ close to $0$, these two distributions should look
similar, and it will be quite difficult to infer whether it is $y$
or $z$. When it comes to the rating vector, it should be $\Pr\left(\mathcal{M}\left(x^{(1)}\right)\in S\right)\le\exp\left(\epsilon\right)\Pr\left(\mathcal{M}\left(x^{(2)}\right)\in S\right)$,
where $x^{(1)}$ and $x^{(2)}$ are two different rating vectors,
which may represent two different users. Hence, the outputs of all
users are mixed up and thus indistinguishable.

\subsection{Modified Laplace mechanism}

In this subsection, we introduce modified Laplace mechanism. The name
comes from Laplace mechanism \cite{dwork2006calibrating}, which works
only on continuous metrizable space. In order to handle missing value,
we will modify it a bit. For convenience, we suppose that the data
are normalized into the interval $\left[-1,1\right]$, and let the
question mark $?$ denote the missing value.
\begin{defn}
For any $\epsilon\geq0$, $\xi\sim\textrm{Laplace}(0,\frac{2}{\epsilon})$
is a random variable, and $\zeta\sim\textrm{Bernoulli}(\frac{\exp(\epsilon/2)}{\exp(\epsilon/2)+1})$
is a random variable independent to $\xi$. A modified Laplace mechanism
$\mathcal{M}\left(x\right)=\left(\mathcal{M}\left(x_{1}\right),\mathcal{M}\left(x_{2}\right),\ldots,\mathcal{M}\left(x_{n}\right)\right)$
is defined as 
\begin{equation}
\mathcal{M}\left(x_{i}\right)=\left\{ \begin{array}{ll}
\zeta\cdot\left(x_{i}+\xi\right)+\left(1-\zeta\right)\cdot? & :x_{i}\in\left[-1,1\right]\\
\zeta\cdot?+\left(1-\zeta\right)\cdot\xi & :x_{i}=?
\end{array}\right.,\label{eq:mlaplace}
\end{equation}
where by convention, $1\cdot?=?$, $0\cdot?=0$ and $?+0=?$.
\end{defn}
The idea is that besides adding Laplace noise, we randomly remove
and create some ratings as well. This mechanism can be proven to be
differentially private.
\begin{thm}
\label{thm:laplace}Modified Laplace mechanism is $n\epsilon$-differentially
private.
\end{thm}

\subsection{Randomized response mechanism}

In this subsection, we present randomized response mechanism, which
works on discrete data. Note that \cite{duchi2013local,kairouz2014extremal}
also use this term, and their definitions are even different between
them. Our use of this term is closer to \cite{kairouz2014extremal},
and we will adapt it to rating data.

Recommender systems rarely allow users to give continuous ratings.
Instead, they often ask the user to rate an item by one to five stars.
These ratings are surely ordinal, but we just ignore the order of
the set. Along with the missing rating, we consider them as cardinal
numbers. In the following definition, the number $0$ can be seen
as the missing rating, and the numbers $1,2,\ldots,d$ can be seen
as the number of stars.
\begin{defn}
Let $W=\left\{ 0,1,2,...,d\right\} $ be a set of finite cardinality.
For any $\epsilon\ge0$, any $i\in W$ , $\xi_{i}$ is a (independent)
random variable with support on $W$ whose probability mass function
is defined by $p_{i}\left(j\right)=\frac{\exp\left(\epsilon\right)^{I\left(j=i\right)}}{\exp\left(\epsilon\right)+d}$
, for any $j\in W$ . A randomized response mechanism $\mathcal{M}\left(x\right)=\left(\mathcal{M}\left(x_{1}\right),\mathcal{M}\left(x_{2}\right),\ldots,\mathcal{M}\left(x_{n}\right)\right)$
is defined as $\mathcal{M}\left(x_{k}\right)=\xi_{x_{k}}$ .
\end{defn}
The idea is that the transformed ratings (including missing ratings)
will most likely remain the same as the original ratings, but there
is still possibility that they are transformed to other ratings (with
equal probability). One can prove that this mechanism is differentially
private.
\begin{thm}
\label{thm:random}Randomized response mechanism is $n\epsilon$-differentially
private.
\end{thm}

\section{Utility\label{sec:Utility}}

A natural question is that whether the transformed ratings are useful.
If the transformed ratings produce nonsense, then there will be no
meaning of this transform although the user privacy is protected.
This question can be decomposed into two subquestions: what the usefulness
means and what could be the possible way to make it useful. For the
first question, we will use the statistical estimation framework.
And for the second one, we will use the low-rank matrix completion
method. 

We start with the framework. There are $m$ users and $n$ items in
the universe. $\Theta_{m\times n}$ is the unknown matrix of true
ratings that each user will give to each item. This is a dense matrix
without any missing values. However, since there are so many items,
users are not able to test every item and their ratings are corrupted
by noise. What we finally observe is a sparse matrix $X_{m\times n}$,
which could be regarded as some approximation of $\Theta$. Then,
we apply either of our mechanism on $X$ to generate the transformed
rating matrix $Z_{m\times n}$. Since our mechanism is computed elementwisely,
this can be done locally at each user's computer. After that, each
user sends their transformed rating vector to the service provider,
who observes the matrix $Z$. The service provider's goal is to recover
$\Theta$ from $Z$.

\begin{figure}[h]
\noindent \begin{centering}
\includegraphics[scale=0.3]{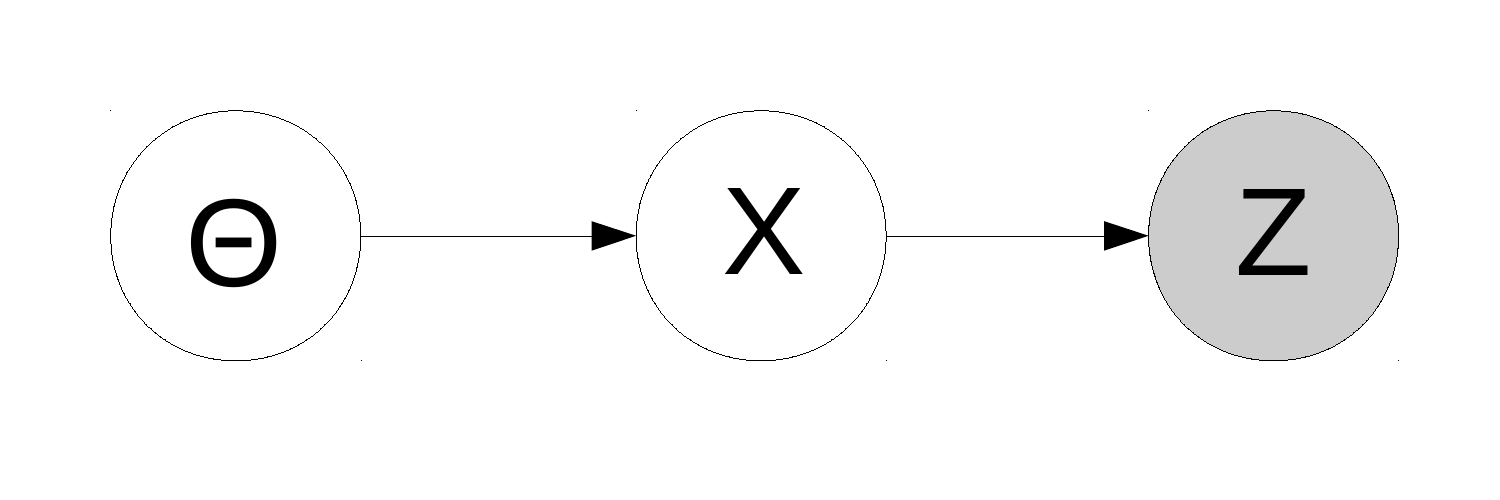}
\par\end{centering}

\protect\caption{Rating generating process}
\end{figure}

Now we will show how it is possible to recover $\Theta$ from $Z$
instead of from $X$. As mentioned in Section~\ref{sec:Introduction},
there are quite a few methods. Interesting readers can refer to \cite{ricci2011introduction,aggarwal2016recommender}.
Here we will only present one method, but the analysis can be generalized
to all methods. 

This method is low-rank matrix completion. We suppose $\Theta$ is
a low-rank matrix, i.e. $\textrm{rank}(\Theta)=r\ll\min(m,n)$. If
the true rating matrix is low-rank, then we are able to approximately
recover it with a few corrupted ratings under certain condition such
as \emph{restricted isometry property} (RIP) \cite{fazel2008compressed}. 
\begin{defn}
Let $\Omega_{Z}$ denote the support of the non-missing ratings of
$Z$. A projection operator $\mathcal{P}_{\Omega_{Z}}$ satisfies
restricted isometry property if it obeys 
\begin{equation}
\left(1-\alpha\right)\left\Vert A\right\Vert _{F}^{2}\leq\tfrac{1}{p}\left\Vert \mathcal{P}_{\Omega_{Z}}\left(A\right)\right\Vert _{F}^{2}\leq\left(1+\alpha\right)\left\Vert A\right\Vert _{F}^{2},\label{eq:rip}
\end{equation}
for any matrix $A$ with sufficiently small rank and $\alpha\in\left(0,1\right)$
sufficiently small, where $p$ is the proportion of non-missing values
of $Z$ and $\left\Vert \cdot\right\Vert _{F}$ is the Frobenius norm.
\end{defn}
The recover process is described as follow. Suppose that 
\begin{equation}
\rho:=\left\Vert \mathcal{P}_{\Omega_{Z}}(\Theta-Z)\right\Vert _{F}<\infty.\label{eq:obserror}
\end{equation}
Our estimator $\hat{\Theta}$ is obtained from the following optimization
problem
\begin{equation}
\begin{aligned}\arg\min_{M} & \qquad\left\Vert M\right\Vert _{*}\\
\text{s.t.} & \quad\left\Vert \mathcal{P}_{\Omega_{Z}}(M-Z)\right\Vert _{F}\le\rho,
\end{aligned}
\label{eq:program}
\end{equation}

where $\left\Vert \cdot\right\Vert _{*}$ is the nuclear norm (a.k.a.
trace norm). 

Under low-rank hypothesis and RIP, \cite{fazel2008compressed} proved
\begin{equation}
\left\Vert \hat{\Theta}-\Theta\right\Vert _{F}\leq C_{0}p^{-1/2}\rho\label{eq:esterror}
\end{equation}
for some numerical constant $C_{0}$. This means that the estimation
error on the whole matrix is proportional to the error on the support
of the observed matrix, which means that the recover method enjoys
a kind of stability against the noise quantified by $\rho$. Of course,
this noise includes not only the noise intrinsic in the problem (i.e.
between $\Theta$ and $X$) but also the noise artificially introduced
through the mechanism (i.e. between $X$ and $Z$). We see how easily
the traditional analysis techniques can be seamlessly transplanted
to the new setting.

What remains is just to give an upper bound of $\rho$. Let $\Omega_{X}$
denote the support of the non-missing ratings of $X$, and $s:=|\Omega_{X}|$
be the number of non-missing ratings. Suppose that $\left\Vert \mathcal{P}_{\Omega_{X}}\left(\Theta-X\right)\right\Vert _{F}\leq\rho_{0}\sqrt{s}<\infty$,
for some small constant $\rho_{0}$. This hypothesis is quite realistic.
Indeed, this is what we need if we want to recover $\Theta$ from
the untransformed ratings $X$. Then we have the following theorems.
\begin{thm}
\label{thm:upper}Let $\gamma\in\left(0,1\right)$ be the level of
tolerance. With probability at least $1-\gamma$, the $Z$ generated
by modified Laplace mechanism satisfies 
\begin{equation}
\rho\leq\rho_{0}\sqrt{s}+\frac{4}{\epsilon}\sqrt{\frac{s}{\gamma}}+\sqrt{\frac{2mn}{(e^{\frac{\epsilon}{2}}+1)\gamma}\left(1+\frac{8}{\epsilon^{2}}\right)};\label{eq:upplaplace}
\end{equation}
with probability at least $1-\gamma$, the $Z$ generated by randomized
response mechanism satisfies

\begin{equation}
\rho\leq\rho_{0}\sqrt{s}+2(d-1)\sqrt{\frac{2mnd}{(e^{\epsilon}+d)\gamma}}.\label{eq:upprandom}
\end{equation}

\end{thm}
When the privacy parameter $\epsilon$ increases toward infinity,
the above upper bounds decrease toward $\rho_{0}\sqrt{s}$. This means
that the less the level of differential privacy is, the more accurate
the data are and then the more precise estimation we will have. This
is intuitive, since the larger $\epsilon$ is, the less extra noise
we introduce into the data. In practice, it is desirable to choose
an $\epsilon$ which makes the entire upper bound match the order
of $\rho_{0}\sqrt{s}$. Combining these bounds with (\ref{eq:esterror}),
we assure the utility of our transformed ratings.

\section{Proof\label{sec:Proof}}

\subsection{Proof of Theorem~\ref{thm:laplace}}
\begin{proof}
According to the values of $(x,y)$ and $S$, we should divide it
into nine cases. We only consider non empty set of $S$, since the
empty set case is trivial. 

i) $(x,y)\in[-1,1]^{2}$ and $S\in\mathcal{R}([-1,1])$
\begin{align*}
\frac{\Pr(\mathcal{M}(x)\in S)}{\Pr(\mathcal{M}(y)\in S)} & =\frac{\Pr(\zeta_{1}=1,x+\xi_{1}\in S)}{\Pr(\zeta_{2}=1,y+\xi_{2}\in S)}\\
 & =\frac{\Pr(\zeta_{1}=1)\Pr(x+\xi_{1}\in S)}{\Pr(\zeta_{2}=1)\Pr(y+\xi_{2}\in S)}\\
 & =\frac{\Pr(x+\xi_{1}\in S)}{\Pr(y+\xi_{2}\in S)}\\
 & \leq e^{\epsilon}.
\end{align*}

ii) $(x,y)\in[-1,1]^{2}$ and $S=\{?\}$
\begin{align*}
\frac{\Pr(\mathcal{M}(x)\in S)}{\Pr(\mathcal{M}(y)\in S)}=\frac{\Pr(\zeta_{1}=0)}{\Pr(\zeta_{2}=0)}=1\leq e^{\epsilon}.
\end{align*}

iii) $(x,y)\in[-1,1]^{2}$ and $S$ is more than $\{?\}$

\begin{align*}
\frac{\Pr(\mathcal{M}(x)\in S)}{\Pr(\mathcal{M}(y)\in S)}=\frac{\Pr(\zeta_{1}=1,x+\xi_{1}\in S)+\Pr(\zeta_{1}=0)}{\Pr(\zeta_{2}=1,y+\xi_{2}\in S)+\Pr(\zeta_{2}=0)}\leq e^{\epsilon}.
\end{align*}

iv) $x\in[-1,1]$, $y=?$, and $S\in\mathcal{R}([-1,1])$

\begin{align*}
\frac{\Pr(\mathcal{M}(x)\in S)}{\Pr(\mathcal{M}(y)\in S)} & =\frac{\Pr(\zeta_{1}=1,x+\xi_{1}\in S)}{\Pr(\zeta_{2}=0,\xi_{2}\in S)}\\
 & =\frac{\Pr(\zeta_{1}=1)\Pr(x+\xi_{1}\in S)}{\Pr(\zeta_{2}=0)\Pr(\xi_{2}\in S)}\\
 & \leq e^{\frac{\epsilon}{2}}\cdot e^{\frac{\epsilon}{2}}\\
 & =e^{\epsilon}.
\end{align*}

v) $x\in[-1,1]$, $y=?$, and $S=\{?\}$

\begin{align*}
\frac{\Pr(\mathcal{M}(x)\in S)}{\Pr(\mathcal{M}(y)\in S)}=\frac{\Pr(\zeta_{1}=0)}{\Pr(\zeta_{2}=1)}=e^{-\frac{\epsilon}{2}}\leq e^{\epsilon}.
\end{align*}

vi) $x\in[-1,1]$, $y=?$ and $S$ is more than $\{?\}$

\begin{align*}
\frac{\Pr(\mathcal{M}(x)\in S)}{\Pr(\mathcal{M}(y)\in S)}=\frac{\Pr(\zeta_{1}=1,x+\xi_{1}\in S)+\Pr(\zeta_{1}=0)}{\Pr(\zeta_{2}=0,\xi_{2}\in S)+\Pr(\zeta_{2}=1)}\leq e^{\epsilon}.
\end{align*}

vii) $y\in[-1,1]$, $x=?$, and $S\in\mathcal{R}([-1,1])$

\begin{align*}
\frac{\Pr(\mathcal{M}(x)\in S)}{\Pr(\mathcal{M}(y)\in S)} & =\frac{\Pr(\zeta_{1}=0,\xi_{1}\in S)}{\Pr(\zeta_{2}=1,y+\xi_{2}\in S)}\\
 & =\frac{\Pr(\zeta_{1}=0)\Pr(\xi_{1}\in S)}{\Pr(\zeta_{2}=1)\Pr(y+\xi_{2}\in S)}\\
 & \leq e^{-\frac{\epsilon}{2}}\cdot e^{\frac{\epsilon}{2}}\\
 & \leq e^{\epsilon}.
\end{align*}

viii) $y\in[-1,1]$, $x=?$, and $S=\{?\}$

\begin{align*}
\frac{\Pr(\mathcal{M}(x)\in S)}{\Pr(\mathcal{M}(y)\in S)}=\frac{\Pr(\zeta_{1}=1)}{\Pr(\zeta_{2}=0)}=e^{\frac{\epsilon}{2}}\leq e^{\epsilon}.
\end{align*}

ix) $y\in[-1,1]$, $x=?$ and $S$ is more than $\{?\}$

\begin{align*}
\frac{\Pr(\mathcal{M}(x)\in S)}{\Pr(\mathcal{M}(y)\in S)}=\frac{\Pr(\zeta_{1}=0,\xi_{1}\in S)+\Pr(\zeta_{1}=1)}{\Pr(\zeta_{2}=1,y+\xi_{2}\in S)+\Pr(\zeta_{2}=0)}\leq e^{\epsilon}.
\end{align*}

\end{proof}

\subsection{Proof of Theorem~\ref{thm:random}}
\begin{proof}
For any $(x,y)\in W^{2}$, we have 
\begin{equation}
\frac{\Pr(\mathcal{M}(x)\in S)}{\Pr(\mathcal{M}(y)\in S)}=\frac{\sum_{s\in S}\Pr(\mathcal{M}(x)=s)}{\sum_{s\in S}\Pr(\mathcal{M}(y)=s)}.\label{eq:category}
\end{equation}
$s$ could be three kinds of values: $x$, $y$ and the other. So
we divide it into three cases. 

i) $s=x$

\[
\frac{\Pr(\mathcal{M}(x)=x)}{\Pr(\mathcal{M}(y)=x)}=\frac{\frac{e^{\epsilon}}{e^{\epsilon}+d}}{\frac{1}{e^{\epsilon}+d}}=e^{\epsilon}.
\]

ii) $s=y$

\[
\frac{\Pr(\mathcal{M}(x)=y)}{\Pr(\mathcal{M}(y)=y)}=\frac{\frac{1}{e^{\epsilon}+d}}{\frac{e^{\epsilon}}{e^{\epsilon}+d}}=e^{-\epsilon}\leq e^{\epsilon}.
\]

iii) $s\neq x$ and $s\neq y$

\[
\frac{\Pr(\mathcal{M}(x)=s)}{\Pr(\mathcal{M}(y)=s)}=\frac{\frac{1}{e^{\epsilon}+d}}{\frac{1}{e^{\epsilon}+d}}=1\leq e^{\epsilon}.
\]
So in either way, the fraction is no more than $e^{\epsilon}$. Join
these equations into (\ref{eq:category}), we get 
\[
\frac{\Pr(\mathcal{M}(x)\in S)}{\Pr(\mathcal{M}(y)\in S)}\leq\frac{\sum_{s\in S}\Pr(\mathcal{M}(y)=s)e^{\epsilon}}{\sum_{s\in S}\Pr(\mathcal{M}(y)=s)}=e^{\epsilon}.
\]

\end{proof}

\subsection{Proof of Theorem~\ref{thm:upper}}
\begin{proof}
Firstly, we decompose \ref{eq:obserror} into three terms. 
\begin{align*}
\left\Vert \mathcal{P}_{\Omega_{Z}}(\Theta-Z)\right\Vert _{F} & =\left\Vert \mathcal{P}_{\Omega_{Z}\cap\Omega_{X}}(\Theta-X)+\mathcal{P}_{\Omega_{Z}\cap\Omega_{X}}(X-Z)+\mathcal{P}_{\Omega_{Z}\setminus\Omega_{X}}(\Theta-Z)\right\Vert _{F}\\
 & \leq\left\Vert \mathcal{P}_{\Omega_{Z}\cap\Omega_{X}}(\Theta-X)\right\Vert _{F}+\left\Vert \mathcal{P}_{\Omega_{Z}\cap\Omega_{X}}(X-Z)\right\Vert _{F}+\left\Vert \mathcal{P}_{\Omega_{Z}\setminus\Omega_{X}}(\Theta-Z)\right\Vert _{F}.
\end{align*}

The first term 
\[
\left\Vert \mathcal{P}_{\Omega_{Z}\cap\Omega_{X}}(\Theta-X)\right\Vert _{F}\le\left\Vert \mathcal{P}_{\Omega_{X}}(\Theta-X)\right\Vert _{F}\le\rho_{0}\sqrt{s}.
\]

Then, we calculate the mathematical expectation of the square of the
second term. 
\begin{equation}
\begin{split}\mathbb{E}\left\Vert \mathcal{P}_{\Omega_{Z}\cap\Omega_{X}}(X-Z)\right\Vert _{F}^{2} & =\mathbb{E}\left[\sum_{(i,j)\in\Omega_{Z}\cap\Omega_{X}}(X_{ij}-Z_{ij})^{2}\right]\\
 & =\mathbb{E}\left[\mathbb{E}\left[\sum_{(i,j)\in\Omega_{Z}\cap\Omega_{X}}(X_{ij}-Z_{ij})^{2}|\Omega_{X},\Omega_{Z}\right]\right]\\
 & =\mathbb{E}\left[\sum_{(i,j)\in\Omega_{Z}\cap\Omega_{X}}\mathbb{E}\left[(X_{ij}-Z_{ij})^{2}|\Omega_{X},\Omega_{Z}\right]\right].
\end{split}
\label{eq:2ndterm}
\end{equation}

In the same way, we also calculate for the third term. 
\begin{equation}
\begin{split}\mathbb{E}\left\Vert \mathcal{P}_{\Omega_{Z}\setminus\Omega_{X}}(\Theta-Z)\right\Vert _{F}^{2} & =\mathbb{E}\left[\sum_{(i,j)\in\Omega_{Z}\setminus\Omega_{X}}(\Theta_{ij}-Z_{ij})^{2}\right]\\
 & =\mathbb{E}\left[\mathbb{E}\left[\sum_{(i,j)\in\Omega_{Z}\setminus\Omega_{X}}(\Theta_{ij}-Z_{ij})^{2}|\Omega_{X},\Omega_{Z}\right]\right]\\
 & =\mathbb{E}\left[\sum_{(i,j)\in\Omega_{Z}\setminus\Omega_{X}}\mathbb{E}\left[(\Theta_{ij}-Z_{ij})^{2}|\Omega_{X},\Omega_{Z}\right]\right].
\end{split}
\label{eq:3rdterm}
\end{equation}

For modified Laplace mechanism

\begin{align*}
\mathbb{E}\left[(X_{ij}-Z_{ij})^{2}|\Omega_{X},\Omega_{Z}\right] & =2\left(\frac{2}{\epsilon}\right)^{2}=\frac{8}{\epsilon^{2}},\quad\forall i,j\in\Omega_{X}\cap\Omega_{Z};\\
\mathbb{E}\left[(\Theta_{ij}-Z_{ij})^{2}|\Omega_{X},\Omega_{Z}\right] & =\Theta_{ij}^{2}+2\left(\frac{2}{\epsilon}\right)^{2}\leq1+\frac{8}{\epsilon^{2}},\quad\forall i,j\in\Omega_{Z}\backslash\Omega_{X}.
\end{align*}

Join these into (\ref{eq:2ndterm}) and (\ref{eq:3rdterm}), we get
\begin{align*}
\mathbb{E}\left\Vert \mathcal{P}_{\Omega_{Z}\cap\Omega_{X}}(X-Z)\right\Vert _{F}^{2} & \leq\frac{8s}{\epsilon^{2}}\\
\mathbb{E}\left\Vert \mathcal{P}_{\Omega_{Z}\setminus\Omega_{X}}(\Theta-Z)\right\Vert _{F}^{2} & \leq\frac{mn}{e^{\frac{\epsilon}{2}}+1}\left(1+\frac{8}{\epsilon^{2}}\right),
\end{align*}
where $s:=|\Omega_{X}|$.

Then, 
\begin{align*}
\Pr\left(\left\Vert \mathcal{P}_{\Omega_{Z}\cap\Omega_{X}}(X-Z)\right\Vert _{F}>\sqrt{\frac{16s}{\epsilon^{2}\gamma}}\right) & \leq\frac{\mathbb{E}\left\Vert \mathcal{P}_{\Omega_{Z}\cap\Omega_{X}}(X-Z)\right\Vert _{F}^{2}}{\frac{16s}{\epsilon^{2}\gamma}}=\frac{\gamma}{2}\\
\Pr\left(\left\Vert \mathcal{P}_{\Omega_{Z}\setminus\Omega_{X}}(\Theta-Z)\right\Vert _{F}>\sqrt{\frac{2mn}{(e^{\frac{\epsilon}{2}}+1)\gamma}\left(1+\frac{8}{\epsilon^{2}}\right)}\right) & \leq\frac{\mathbb{E}\left\Vert \mathcal{P}_{\Omega_{Z}\setminus\Omega_{X}}(\Theta-Z)\right\Vert _{F}^{2}}{\frac{2mn}{(e^{\frac{\epsilon}{2}}+1)\gamma}\left(1+\frac{8}{\epsilon^{2}}\right)}=\frac{\gamma}{2}.
\end{align*}

Finally, we have 
\[
\Pr\left(\left\Vert \mathcal{P}_{\Omega_{Z}}(\Theta-Z)\right\Vert _{F}\leq\rho_{0}\sqrt{s}+\frac{4}{\epsilon}\sqrt{\frac{s}{\gamma}}+\sqrt{\frac{2mn}{(e^{\frac{\epsilon}{2}}+1)\gamma}\left(1+\frac{8}{\epsilon^{2}}\right)}\right)\geq1-\gamma.
\]

For randomized response mechanism

\begin{align*}
\mathbb{E}\left[(X_{ij}-Z_{ij})^{2}|\Omega_{X},\Omega_{Z}\right] & \leq(d-1)^{2}\frac{d}{e^{\epsilon}+d},\quad\forall i,j\in\Omega_{X}\cap\Omega_{Z};\\
\mathbb{E}\left[(\Theta_{ij}-Z_{ij})^{2}|\Omega_{X},\Omega_{Z}\right] & =(d-1)^{2},\quad\forall i,j\in\Omega_{Z}\backslash\Omega_{X}.
\end{align*}

Join these into (\ref{eq:2ndterm}) and (\ref{eq:3rdterm}), we get
\begin{align*}
\mathbb{E}\left\Vert \mathcal{P}_{\Omega_{Z}\cap\Omega_{X}}(X-Z)\right\Vert _{F}^{2} & \leq(d-1)^{2}\frac{sd}{e^{\epsilon}+d},\\
\mathbb{E}\left\Vert \mathcal{P}_{\Omega_{Z}\setminus\Omega_{X}}(\Theta-Z)\right\Vert _{F}^{2} & \leq\frac{mnd}{e^{\epsilon}+d}(d-1)^{2},
\end{align*}
where $s:=|\Omega_{X}|$.

Then, 
\begin{align*}
\Pr\left(\left\Vert \mathcal{P}_{\Omega_{Z}\cap\Omega_{X}}(X-Z)\right\Vert _{F}>\sqrt{\frac{2sd(d-1)^{2}}{(e^{\epsilon}+d)\gamma}}\right) & \leq\frac{\mathbb{E}\left\Vert \mathcal{P}_{\Omega_{Z}\cap\Omega_{X}}(X-Z)\right\Vert _{F}^{2}}{\frac{2sd(d-1)^{2}}{(e^{\epsilon}+d)\gamma}}=\frac{\gamma}{2}\\
\Pr\left(\left\Vert \mathcal{P}_{\Omega_{Z}\setminus\Omega_{X}}(\Theta-Z)\right\Vert _{F}>\sqrt{\frac{2mnd(d-1)^{2}}{(e^{\epsilon}+d)\gamma}}\right) & \leq\frac{\mathbb{E}\left\Vert \mathcal{P}_{\Omega_{Z}\setminus\Omega_{X}}(\Theta-Z)\right\Vert _{F}^{2}}{\frac{2mnd(d-1)^{2}}{(e^{\epsilon}+d)\gamma}}=\frac{\gamma}{2}.
\end{align*}

Finally, since $s\leq mn$, we have 
\[
\Pr\left(\left\Vert \mathcal{P}_{\Omega_{Z}}(\Theta-Z)\right\Vert _{F}\leq\rho_{0}\sqrt{s}+2(d-1)\sqrt{\frac{2mnd}{(e^{\epsilon}+d)\gamma}}\right)\geq1-\gamma.
\]

\end{proof}
\bibliographystyle{unsrt}
\bibliography{DPRCM}

\end{document}